\documentclass[10pt,letterpaper]{article}
\usepackage[top=0.85in,left=2.75in,footskip=0.75in]{geometry}

\usepackage{amsmath,amssymb}

\usepackage{changepage}

\usepackage{textcomp,marvosym}

\usepackage{cite}

\usepackage{nameref,hyperref}

\usepackage[right]{lineno}

\usepackage[nopatch=eqnum]{microtype}
\DisableLigatures[f]{encoding = *, family = * }

\usepackage[table]{xcolor}

\usepackage{array}

\usepackage{times}
\usepackage{soul}
\usepackage{url}
\usepackage{booktabs}
\usepackage{multirow}
\usepackage{color}
\usepackage{subfigure}
\usepackage{amsmath}
\usepackage{amsthm}

\usepackage{algorithm}
\usepackage{algorithmic}
\newcolumntype{+}{!{\vrule width 2pt}}

\newlength\savedwidth

\raggedright
\usepackage[aboveskip=1pt,labelfont=bf,labelsep=period,justification=raggedright,singlelinecheck=off]{caption}

\makeatletter
\renewcommand{\@biblabel}[1]{\quad#1.}
\makeatother

\usepackage{lastpage,fancyhdr,graphicx}
\usepackage{epstopdf}
\fancyheadoffset[L]{2.25in}
\fancyfootoffset[L]{2.25in}
\begin{document}
\vspace*{0.2in}

\begin{flushleft}
  {\Large
    \textbf\newline{Knowledge-aware contrastive heterogeneous molecular graph learning}
  }
  \newline
  \\
  Mukun Chen\textsuperscript{1},
  Jia Wu\textsuperscript{2},
  Shirui Pan\textsuperscript{3},
  Fu Lin\textsuperscript{1},
  Bo Du\textsuperscript{1},
  Xiuwen Gong\textsuperscript{4},
  Wenbin Hu\textsuperscript{1,*}
  \\
  \bigskip
  \textbf{1} School of Computer Science, Wuhan University, Wuhan, Hubei Province, China
  \\
  \textbf{2} School of Computing, Macquarie University, Sydney, Australia
  \\
  \textbf{3} School of Information and Communication Technology, Griffith University, Queensland, Australia
  \\
  \textbf{4} Faculty of Engineering and Information Technology, University of Technology Sydney, Sydney, Australia
  \\
  \bigskip

  * hwb@whu.edu.cn

\end{flushleft}
\section*{Abstract}
Molecular representation learning is pivotal in predicting molecular properties and advancing drug design. Traditional methodologies, which predominantly rely on homogeneous graph encoding, are limited by their inability to integrate external knowledge and represent molecular structures across different levels of granularity.
To address these limitations, we propose a paradigm shift by encoding molecular graphs into heterogeneous structures, introducing a novel framework: Knowledge-aware Contrastive Heterogeneous Molecular Graph Learning. This approach leverages contrastive learning to enrich molecular representations with embedded external knowledge.
KCHML conceptualizes molecules through three distinct graph views—molecular, elemental, and pharmacological—enhanced by heterogeneous molecular graphs and a dual message-passing mechanism. This design offers a comprehensive representation for property prediction, as well as for downstream tasks such as drug-drug interaction prediction.
Extensive benchmarking demonstrates KCHML's superiority over state-of-the-art molecular property prediction models, underscoring its ability to capture intricate molecular features.

\section*{Author summary}
In the field of drug discovery, predicting molecular properties and drug interactions is crucial for developing new medications and ensuring patient safety. Traditional methods for representing molecular structures often fail to incorporate external knowledge and struggle to capture complex interactions at different levels of detail. To address these limitations, we developed a new framework called Knowledge-aware Contrastive Heterogeneous Molecular Graph Learning (KCHML).

Our approach integrates information from three perspectives—molecular structure, elemental relationships, and pharmacological data—using advanced machine learning techniques. This combination allows for a more detailed and accurate representation of molecules, leading to better predictions of molecular properties and drug interactions. By improving how we model and understand molecules, our work has the potential to streamline drug development and reduce the risk of harmful drug interactions, contributing to safer and more effective treatments.


\section*{Introduction}

At the core of computational drug discovery lies Molecular Representation Learning (MRL), a field that integrates state-of-the-art machine learning techniques with biomedical applications. MRL not only enhances our understanding of molecular interactions but also refines predictive models that are critical for both drug property and drug-drug interaction (DDI) prediction, two key tasks in biological research. By transforming extensive molecular datasets into actionable insights, MRL empowers researchers to explore novel therapeutic avenues and tailor treatments to specific biological markers, while also predicting potential drug interactions that could affect treatment outcomes. The accuracy of these predictions is crucial for the evaluation and selection of molecules across a wide range of applications, from therapeutic interventions to industrial chemicals. This precision facilitates the early identification of promising candidates, streamlining the drug development process, mitigating the risk of costly late-stage failures, and ensuring the safe combination of drugs.

MRL involves the rigorous study of molecular structures and encoding strategies, with advanced models adept at capturing the complexities of molecular geometries, bond types, and functional groups—key factors in both the precise prediction of chemical properties and the prediction of drug interactions \cite{ijcai2024p234,ijcai2024p235}. Various graph neural network architectures, such as GCN \cite{kipf2016semi}, GIN \cite{xu2018powerful}, GAT \cite{velickovic2017graph}, GGNN \cite{li2016gated}, and GraphSage \cite{hamilton2017inductive}, offer distinct approaches to molecular structure learning. Increasingly, MRL has been aligned with the Message Passing Neural Network (MPNN) framework, as established by Gilmer et al. \cite{gilmer2017neural}, which has emerged as a fundamental paradigm in the field. These models emphasize the graph-based topologies of molecular structures, with advanced variants such as DMPNN \cite{yang2019analyzing}, CMPNN \cite{song2020communicative}, and CoMPT \cite{ijcai2021p0309} leveraging both node and edge attributes to enhance message-passing efficiency and accuracy for tasks such as molecular property prediction and DDI prediction.

Recent advancements in self-supervised learning, exemplified by context prediction and attribute masking in PreGNN \cite{hu2020strategies} and GROVER \cite{rong2020self}, have shown remarkable potential in both molecular property and DDI prediction. These approaches introduce advanced methodologies for learning molecular representations but remain primarily focused on local structural properties, often neglecting the integration of external knowledge such as drug-target interactions or therapeutic outcomes. To address this gap, knowledge graph (KG)-based methods, including KGNN \cite{lin2020kgnn}, MDNN \cite{lyu2021mdnn} and MKG-FENN \cite{wu2024mkg}, have emerged, framing molecules as interconnected nodes to incorporate external pharmacological insights. These models provide a more comprehensive perspective, blending molecular and therapeutic views. However, a critical limitation of current KG-based frameworks lies in their inability to seamlessly integrate the detailed microscopic features of drug molecules with their broader biological roles, particularly when predicting interactions between drugs.

Despite significant strides, molecular representation learning continues to face profound challenges. The inherent complexity and heterogeneity of molecular structures frequently hinder the formation of robust embedded representations. Furthermore, traditional methods relying on homogeneous graph encoding are constrained by their limited capacity to incorporate external knowledge and often fail to capture the multi-granular intricacies of molecular structures. Consequently, integrating structural and pharmacological data into a cohesive model remains a complex yet essential task.

\begin{figure}[htbp]
  \centerline{\includegraphics[width=0.9\textwidth]{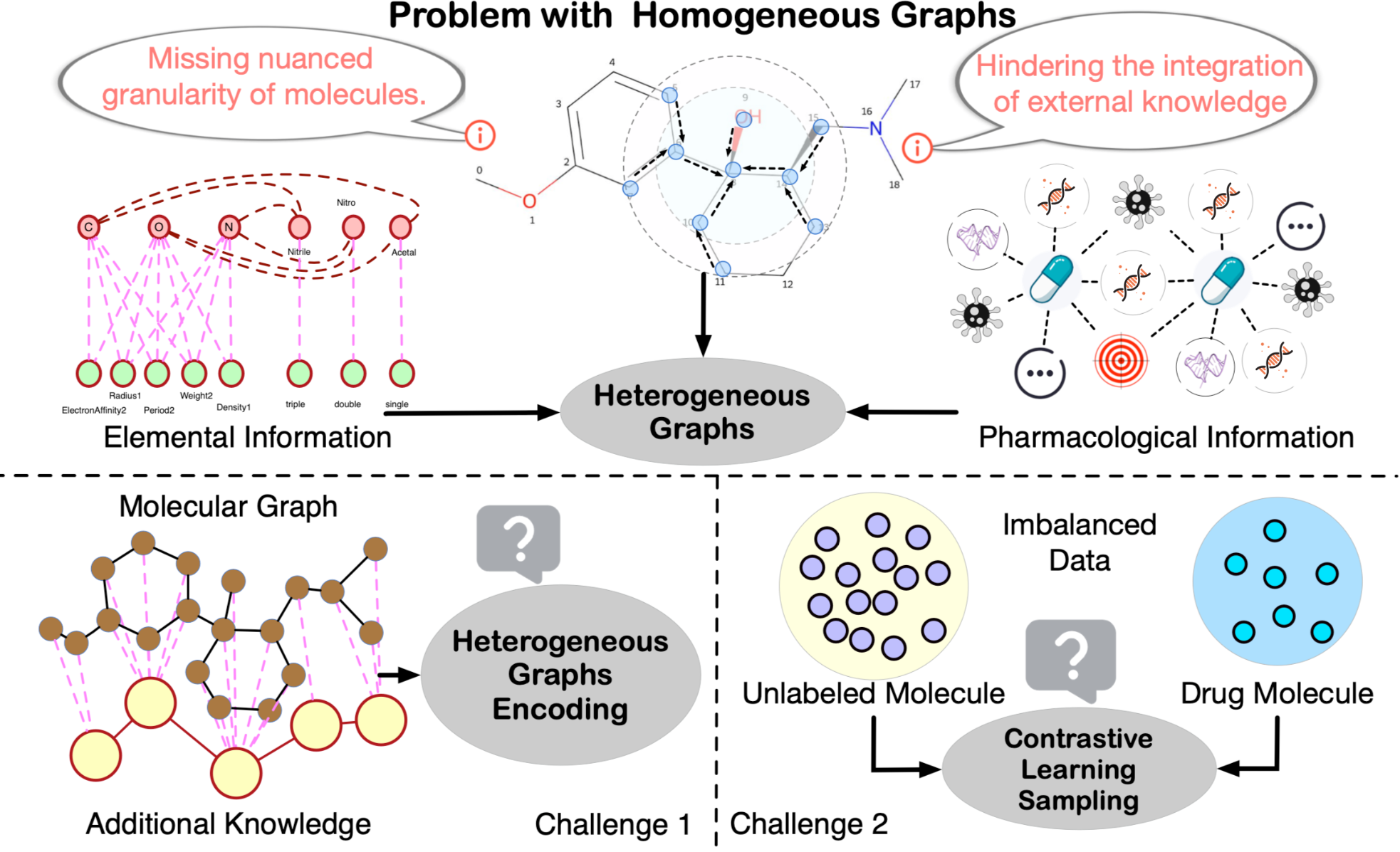}}
  \caption{Homogeneous graphs struggle to capture the nuanced granularity of molecules and hinder the integration of external knowledge. Introducing external knowledge through heterogeneous graphs mainly involves two challenges: (1) heterogeneous graph encoding and (2) imbalanced knowledge sampling.}
  \label{limitation}
\end{figure}

As the field progresses, contrastive learning has emerged as a powerful technique, enhancing the generalizability and resilience of graph encoders. However, significant challenges persist, as illustrated in Fig.~\ref{limitation}. Graph augmentation strategies—such as node dropping, edge perturbation, attribute masking, and subgraph generation \cite{you2020graph}—can unintentionally compromise the fidelity of molecular structures, particularly when external knowledge is integrated \cite{sun2021mocl}. While these augmentations offer new views of molecular configurations, they often overlook the profound impact of minor structural perturbations on pharmacological properties and drug interactions. For example, perturbing edges within a benzene ring, a fundamental structural motif known for its stability and distinctive chemical reactivity, can misrepresent the molecule’s aromatic characteristics, potentially misleading the model. Similarly, removing specific nodes, such as a chlorine atom, risks eliminating crucial information regarding the molecule’s reactivity and solubility, given chlorine's critical role in the biological activity of many pharmaceutical compounds. Most models fall short in accounting for these subtleties during contrastive learning sample generation, especially when predicting drug interactions.

Meanwhile, sample sizes vary across different knowledge levels; elemental knowledge applies universally to molecules, whereas drug knowledge pertains only to pharmacologically active molecules that have been identified as drugs. Drug knowledge encompasses comprehensive data on drug efficacy, mechanisms of action, metabolic pathways, and potential side effects—details not relevant to non-drug molecules. This creates a challenge in sample distribution, as non-drug molecules vastly outnumber drug molecules, leading to potential imbalances in training. Such imbalances can cause the model to overemphasize elemental data while underutilizing the detailed pharmaceutical insights that are critical for predicting complex drug properties and interactions.

To address these challenges, we introduce the cross-view Knowledge-aware Contrastive Heterogeneous Molecular Graph Learning (KCHML) methodology. KCHML seeks to surpass existing methods by integrating structural, functional, and pharmacological perspectives of molecules. It employs a tripartite view framework: the molecular view, the element view, and the drug view. This framework integrates molecular structure, elemental knowledge, and pharmacological data, which are processed through a dual message-passing mechanism. By leveraging contrastive learning, KCHML constructs both positive and negative examples across views, enhancing the model’s ability to discern subtle molecular differences. This enables improved predictions of complex molecular properties and drug interactions, even with the smaller sample size of drug-specific data. This study makes the following contributions:

\begin{itemize}
  \item We present KCHML, a tripartite view framework for molecular property prediction and DDI prediction, integrating molecular, element, and drug views to offer distinct insights into molecular characteristics at different levels.

  \item An innovative encoder has been introduced to manage HMGs, leveraging a dual message-passing mechanism tailored to variations in connectivity and feature distribution across node and edge types.

  \item We utilize a cross-view contrastive learning strategy, focusing on molecular semantics from three perspectives and examining the relationship between contrastive loss and mutual information across views, improving learning efficacy for both molecular property and drug interaction prediction.

  \item Extensive experiments demonstrate KCHML's robust performance in both molecular property prediction and DDI prediction, showcasing its comprehensive understanding of molecules across diverse tasks by integrating elemental, structural, and pharmacological perspectives.
\end{itemize}

\section*{Materials and methods}
\subsection*{Preliminaries}
In this paper, scalars are denoted using lowercase letters (e.g., $x$), vectors by bold lowercase (e.g., $\mathbf{x}$), and matrices with bold uppercase letters (e.g., $\mathbf{X}$). Sets are represented in uppercase italics (e.g., $\mathcal{X}$).

\subsubsection*{Elemental Knowledge Graphs}
The elemental KG, denoted as $\mathcal{G}^{E} = \{(h, r, t) \mid h, t \in \mathcal{V}^{E}, r \in \mathcal{R}^{E}\}$, is structured hierarchically to represent various levels of chemical knowledge. In this context, $\mathcal{V}^{E}$ comprises entities, while $\mathcal{R}^{E}$ captures the relationships between them. The KG is divided into three distinct levels, each offering progressively finer granularity of information.

At the highest level, "class" nodes convey broad categorical concepts, defining high-level chemical classifications and relationships. For example, the triple (``ReactiveNonmetal", ``isSubClassOf", ``Nonmetals") establishes a hierarchical link between "ReactiveNonmetal" and its parent class "Nonmetals," encapsulating abstract chemical groupings.

The intermediate level encompasses core chemical entities such as elements (e.g., "C", "N", "O") and functional groups (e.g., "Nitrile", "Nitro", "Acetal"). Triples at this level, like (``C", ``isPartOf", ``Acetal"), indicate that carbon is part of the functional group "Acetal," denoted in SMARTS notation as ``O[CH1][OX2H0]". These relationships define how elements combine to form higher-order chemical structures.

At the most granular level, property nodes capture specific attributes of chemical entities, such as atomic weight or periodicity. An example triple (``C", ``hasWeight", ``Weight2") denotes that carbon’s atomic weight falls within the "Weight2" category, while (``O", ``isInPeriod", ``Period2") places oxygen within Period 2 of the periodic table.

This hierarchical framework enables the elemental KG to integrate conceptual, structural, and property-level information, offering a comprehensive representation of chemical knowledge from broad classifications to specific properties.

\subsubsection*{Drug Knowledge Graph}

The drug KG, denoted as $\mathcal{G}^{D} = {(h, r, t) \mid h, t \in \mathcal{V}^{D}, r \in \mathcal{R}^{D}}$, is an integrated biological network that encompasses entities categorized as drugs, along with associated concepts such as genes, compounds, diseases, biological processes, side effects, and symptoms. This expansive structure organizes these entities while intricately mapping the complex interactions and relationships that exist among them.

In $\mathcal{G}^{D}$, $\mathcal{V}^{D}$ represents the entities, which range from drug molecules to biological markers, and $\mathcal{R}^{D}$ defines the various types of relationships, such as drug-gene interactions, drug-disease associations, and drug-side effect linkages. The drug KG thus provides a comprehensive knowledge framework, capturing the diverse roles that drugs play within biological and medical systems, and offering deep insights into their interactions with biological pathways and molecular targets.

\subsubsection*{Heterogeneous Molecular Graph}
Based on the integrated information sources, we categorize the Heterogeneous Molecular Graph (HMG) into three distinct views:
\begin{itemize}
  \item Molecule View $\mathcal{G}^{M}$: Generated solely by RDKit using the Simplified Molecular Input Line Entry System (SMILES), this view provides a foundational representation of molecular structures, focusing on atoms, bonds, and connectivity. It does not require external knowledge, capturing basic molecular details.
  \item Element View $\mathcal{G}^{EM}$: This view enhances the molecule view $\mathcal{G}^{M}$ by integrating nodes from the element knowledge graph $\mathcal{G}^{E}$. It incorporates chemical domain knowledge, such as elemental properties and functional groups, thereby enriching the graph’s connectivity. This added information helps model molecular interactions with greater detail by incorporating atomic-level chemical data.
  \item Drug View $\mathcal{G}^{DM}$: Augmented by the drug KG $\mathcal{G}^{D}$, this view includes a Drug Node (DNode) that acts as a central hub linking various nodes related to drugs, including genes, biological processes, and diseases. Initialized with embeddings from the drug KG, it integrates extensive pharmacological knowledge, such as drug efficacy, side effects, and biological mechanisms. This view is essential for guiding molecular pre-training by providing external insights and established drug properties, thus facilitating more accurate downstream predictions, sunch as DDI.
\end{itemize}

\subsubsection*{Problem Formulation}
This study aims to develop a self-supervised graph encoder, denoted as \( h = f(\mathcal{G}^{M}) \in \mathbb{R}^d \), that transforms molecular graphs into high-dimensional vectors without relying on external labels. The training of this encoder leverages knowledge graphs \(\mathcal{G}^{E}\) and \(\mathcal{G}^{D}\) to enrich the encoding process with contextual information.

\subsection*{Framework Overview}
\begin{figure*}[!ht]
  \begin{adjustwidth}{-1.5in}{0in}
    \centering\includegraphics[width=1.25\textwidth]{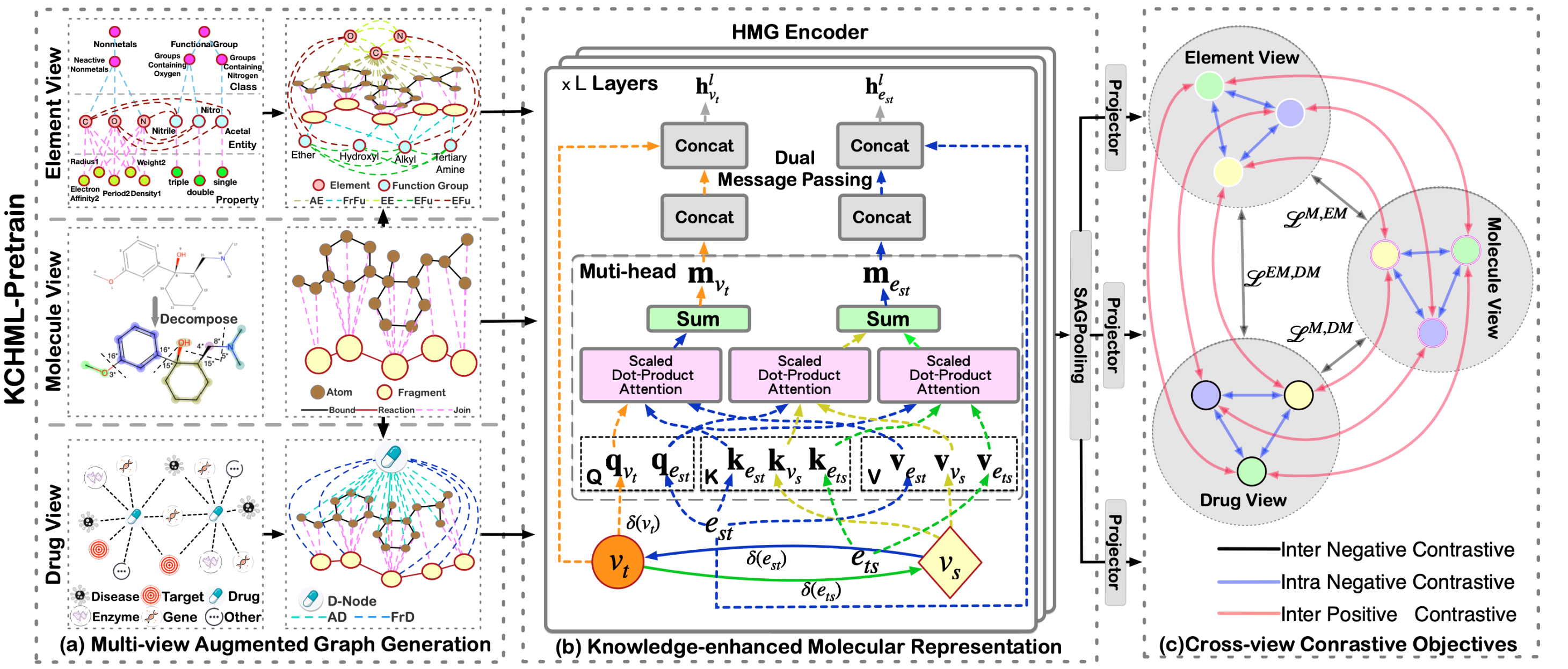}
    \caption{Illustration of the KCHML model. \textbf{(a)} illustrates the three views of HMG, based on the molecular view, the element view is formed by adding two types of nodes and five types of edges, and the drug view is formed by adding one type of node and two types of edges. \textbf{(b)} describes the encoding process of the HMG encoder in detail. The lines of different colors indicate the source of the ($\mathbf{Q}, \mathbf{K}, \mathbf{V}$) of different nodes and edges. For example, the message of node $v_t$ is formed by edge $e_{st}$, and the message of edge $e_{st}$ is provided by $v_s$ and $e_{ts}$. \textbf{(c)} describes the construction process of contrastive learning sample pairs across multiple views.}
    \label{framework}
  \end{adjustwidth}
\end{figure*}

As illustrated in Fig.\ref{framework}, the cross-view KCHML approach consists of three key components: (1) multi-view augmented graph generation, (2) knowledge-enhanced molecular representation, and (3) cross-view contrastive objectives. This section provides an overview of each component.

\subsection*{Multi-view Augmented Graph Generation}
\paragraph*{Molecule View}
Within the molecule view, we delineate two node types and three edge types:
\begin{itemize}
  \item \textbf{Atom-Bond-Atom}: These edges denote the chemical bonds linking atoms within the molecule, representing its foundational chemical structure.
  \item \textbf{Fragment-Reaction-Fragment}: Generated through the BRICS molecular fragmentation algorithm \cite{BRICS}, where "Fragment" signifies molecular segments and "Reaction" denotes the breakpoints between these fragments.
  \item \textbf{Atom-Join-Fragment}: Represents the association between an atom and its corresponding fragment within the molecule.
\end{itemize}

\paragraph*{Element View}:
Expanding from the molecule view \(\mathcal{G}^{M}\), we incorporate two additional node types and introduce five new edge types:

\begin{itemize}
  \item \textbf{Atom-AE-Element}: Links an atom node in \(\mathcal{G}^{M}\) to an element node in \(\mathcal{G}^{E}\) based on shared chemical symbols (e.g., linking a "C" atom to a "C" element).
  \item \textbf{Fragment-FrFu-Functional Group}: Formed by detecting functional groups within molecular fragments in \(\mathcal{G}^{M}\) that correspond to entries in \(\mathcal{G}^{E}\).
  \item \textbf{Element-EE-Element}: Represents multi-hop connections between element nodes in \(\mathcal{G}^{E}\), retaining only 2-hop connections in the Heterogeneous Molecular Graph (HMG). Edge features are determined by the attributes of intermediate nodes traversed.
  \item \textbf{Functional Group-FuFu-Functional Group}: Depicts multi-hop connections between functional group nodes in \(\mathcal{G}^{E}\), with 2-hop connections preserved in the HMG. Edge features are defined by intermediate nodes traversed.
  \item \textbf{Element-EFu-Functional Group}: Transfers relationships from \(\mathcal{G}^{E}\) between elements and functional groups into the HMG.
\end{itemize}

Note that multiple common attributes between elements and functional groups in \(\mathcal{G}^{E}\) result in the addition of new edges representing EE, FuFu, or EFu. This allows for the presence of multiple edges between any pair of nodes in \(\mathcal{G}^{EM}\).

\paragraph*{Drug View}:
Integrating the drug KG \(\mathcal{G}^{D}\) into the molecular view introduces a new type of node and two types of edges, linking each node to the DNode:
\begin{itemize}
  \item \textbf{Atom-AD-DNode}: Connects each atom node in \(\mathcal{G}^{M}\) to DNode.
  \item \textbf{Fragment-FrD-DNode}: Connects each fragment node in \(\mathcal{G}^{M}\) to DNode.
\end{itemize}

Notably, not all molecules have corresponding drug IDs. Therefore, we have devised batch generation strategies and cross-view contrastive learning techniques to handle these scenarios effectively.

\subsection*{Knowledge-Enhanced Molecular Representation}
The encoding of the Heterogeneous Molecular Graph (HMG) focuses on implementing sophisticated message-passing mechanisms crucial for propagating node features across the network. Inspired by the Graph Transformer architecture \cite{maziarka2020molecule}, our approach adeptly navigates through diverse node and edge types within the HMG. This iterative process involves updating node states by aggregating neighborhood features, thereby capturing both local details and global molecular characteristics comprehensively.

Algorithm \ref{alg1} outlines the detailed procedure for encoding the HMG, emphasizing the adaptability of our model to incorporate various node and edge types effectively. Within the element view, a dual message-passing mechanism manages multiple edge connections efficiently, ensuring each type contributes distinct information that enhances the semantic robustness of the molecular representation.

\begin{algorithm}[!ht]
  \caption{HMG encoding algorithm.}
  \label{alg1}
  \begin{algorithmic}[1]
    \REQUIRE The HMG $\mathcal{G}=\{\mathcal{V},\mathcal{E}\}$;
    depth $L$; input node features $\{\mathbf{\alpha}_i\}$; input edge features $\{\mathbf{\beta}_{ij}\}$.
    \ENSURE Graph embedding $\mathbf{h}_{\mathcal{G}}$.
    \STATE  $\mathbf{h}_{v_i}^0 = Init(\alpha_i),\forall {v_i} \in \mathcal{V}$; $\mathbf{h}^0_{{e}_{ij}}  = Init(\mathbf{\beta}_{ij}), \forall {{e}_{ij}} \in \mathcal{E}$
    \FOR{\( l = 1 \) to \( L \)}
    \FOR{\( v_i \in \mathcal{V} \)}
    \STATE $\mathbf{m}^{l}_{v_i} = \text{AGG}^{(l)}(\{e^{l-1}_{pi}|{v_p} \in \mathcal{N}(v_i)\})$
    \STATE  $\mathbf{h}^{l}_{v_{i}} = \text{UPDATE}(\mathbf{h}^{l-1}_{v_i}, \mathbf{m}^{l}_{v_i})$
    \ENDFOR
    \FOR{\( e_{ij} \in  \mathcal{E} \)}
    \STATE $\mathbf{m}^{l}_{e_{ij}}  =  \text{AGG}^{(l)}(\{v_i\}\cup \{e^{l-1}_{pi} : {v_p} \in \mathcal{N}(v_i)\})$
    \STATE $\mathbf{h}^{l}_{e_{ij}} = \text{UPDATE}(\mathbf{h}^{l-1}_{e_{ij}}, \mathbf{m}^{l}_{e_{ij}})$
    \ENDFOR
    \ENDFOR
    \STATE $\mathbf{h}_{\mathcal{G}} = \text{READOUT}(\{\mathbf{h}^{L}_{v_{i}}\},(\{\mathbf{h}^{L}_{e_{ij}}\}))$
  \end{algorithmic}
\end{algorithm}

\begin{figure}[!ht]
  \centerline{\includegraphics[width=0.98\textwidth]{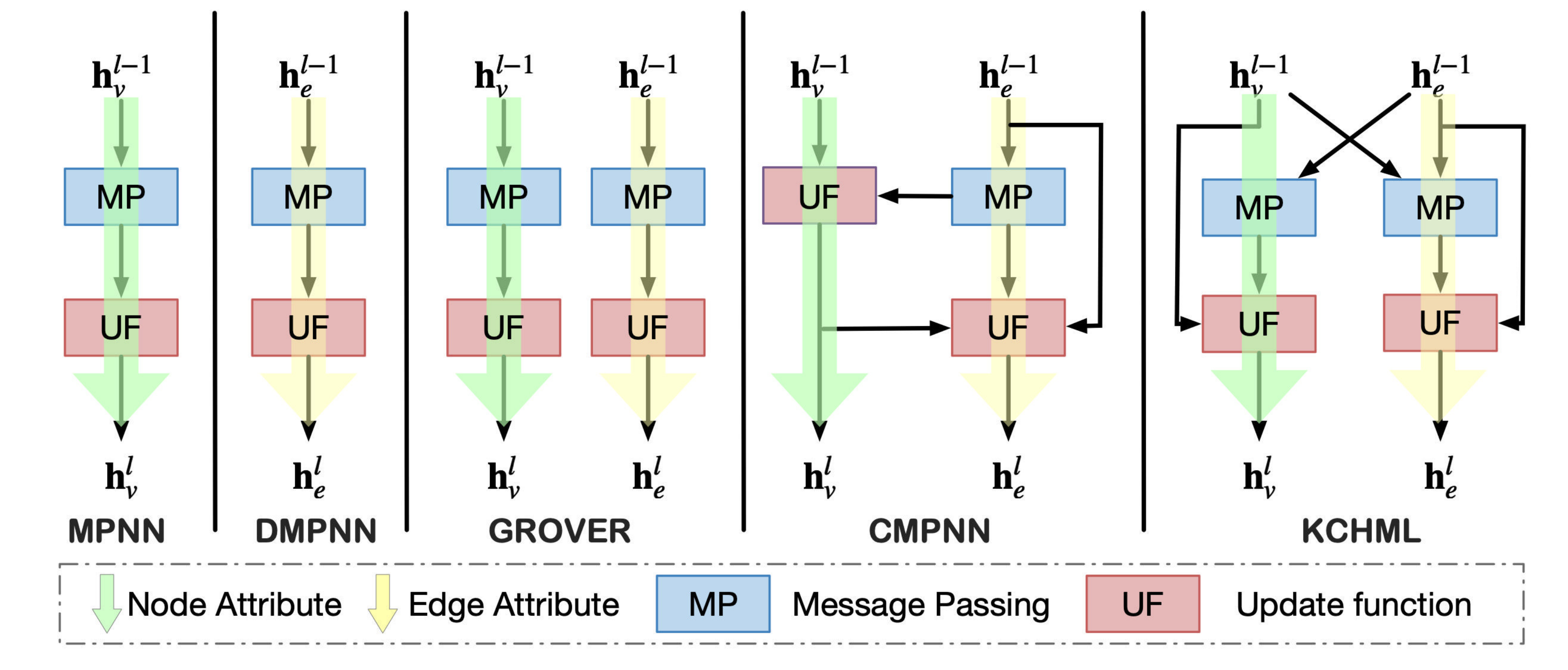}}
  \caption{Comparison of MPNN, DMPNN, GROVER, CMPNN, and KCHML}
  \label{compare}
\end{figure}

Fig.\ref{compare} depicts the message-passing mechanisms of various prominent molecular graph encoders. The primary distinctions of KCHML compared to other encoders include (1) the utilization of distinct mapping spaces for different types of edges and nodes, and (2) the cross-transmission of edge and node information via dual branches within the encoder.

\subsubsection*{Initialize Input}
For a given node \( v_i \in \mathcal{V} \) with features \( \mathbf{\alpha}_i \in \mathbb{R}^{d_v} \) and an edge \( e_{ij} \in \mathcal{E} \) with features \( \mathbf{\beta}_{ij} \in \mathbb{R}^{d_e} \), these input features \( \mathbf{\alpha}_i \) and \( \mathbf{\beta}_{ij} \) undergo linear projection to be embedded into \( d \)-dimensional hidden features \( \mathbf{h}^0_{v_i} \) and \( \mathbf{h}^0_{e_{ij}} \), respectively.
When encoding the edge \( e_{ij} \), we integrate the encoding of the source node \( v_i \) into its initial representation. It's important to note that \( \mathbf{h}_{e_{ij}} \neq \mathbf{h}_{e_{ji}} \), indicating that distinct representations are maintained for each direction of the edge between connected nodes.

\begin{equation}
  \begin{split}
    \mathbf{h}_{v_i}^0      & = \text{Init}(\alpha_i) = (\alpha_i\mathbf{A}_{\delta(v_i)} + \mathbf{a}^0) + ( \mathbf{\lambda}_i\mathbf{C}^0 + \mathbf{c}^0); \\
    \mathbf{h}^0_{{e}_{ij}} & = \text{Init}(\mathbf{\beta}_{ij}) = \mathbf{\beta}_{ij}\mathbf{B}_{\delta(e_{ij})} + \mathbf{b}^0 + \mathbf{h}_{v_i}^0
  \end{split}
\end{equation}
where \( \mathbf{A}_{\delta(v_i)} \in \mathbb{R}^{d_v \times d} \) and \( \mathbf{B}_{\delta(e_{ij})} \in \mathbb{R}^{d_e \times d} \) represent type-specific parameters for the linear projection layers corresponding to nodes and edges, respectively. The bias terms are denoted as \( \mathbf{a}^0, \mathbf{b}^0 \in \mathbb{R}^d \). The functions \( \delta(v_i) \) and \( \delta(e_{ij}) \) determine the specific mapping spaces used for different node and edge types. Additionally, \( \mathbf{C}^0 \in \mathbb{R}^{k \times d} \) and \( \mathbf{c}^0 \in \mathbb{R}^d \) are employed to encode the positional information \( \mathbf{\lambda}_i \in \mathbb{R}^{k} \) of the node \( v_i \).

\subsubsection*{Dual Message Passing Mechanisms}

The KCHML framework incorporates a dual message-passing mechanism that operates on the principle of simultaneous propagation between nodes and edges. This approach enables nodes to receive messages from their adjacent edges while allowing edges to aggregate information from both their source nodes and neighboring edges. This bi-directional exchange of information across the HMGs enhances the interconnectedness and richness of the representations.
The dual message-passing mechanism is pivotal in facilitating the simultaneous propagation of information between nodes and edges, thereby creating more comprehensive and interconnected representations within the KCHML framework.

Our approach diverges from the standard Transformer architecture in two fundamental ways. Firstly, when computing queries, keys, and values (\( \mathbf{Q}, \mathbf{K}, \mathbf{V} \)), we employ linear mappings that project heterogeneous node and edge types into a unified feature space. This transformation ensures that despite their inherent differences, all nodes and edges can be processed within a shared space. This unified space facilitates more effective aggregation and comparison of features across different types. Specifically:
\begin{equation}
  \begin{split}
    \mathbf{q}^{l,k} & =\mathbf{h}^{l-1} \mathbf{W}^{l,k}_{Q} \mathbf{W}_{\delta(v)} ; \\
    \mathbf{k}^{l,k} & =\mathbf{h}^{l-1} \mathbf{W}^{l,k}_{K} \mathbf{W}_{\delta(v)} ; \\
    \mathbf{v}^{l,k} & =\mathbf{h}^{l-1} \mathbf{W}^{l,k}_{V} \mathbf{W}_{\delta(v)}
  \end{split}
\end{equation}
Here, \( \mathbf{W}^{l,k}_{Q} \), \( \mathbf{W}^{l,k}_{K} \), and \( \mathbf{W}^{l,k}_{V} \) are linear projection matrices designed to map the input embeddings into multiple attention heads. The matrix \( \mathbf{W}_{\delta(v)} \) represents a type-specific transformation that ensures all node types are projected consistently into a unified feature space. This process enables the model to generate uniform attention representations across diverse node types. The term \( l = 1 \cdots L \) denotes the current layer index in the stack, while \( k = 1 \cdots K \) indicates the number of attention heads. Similarly, edge features undergo a similar transformation, where the edge type detection function \( \delta(e) \) is used to project edge types into a shared feature space.

Secondly, rather than using a conventional self-attention mechanism, we introduce a specialized attention mechanism. In this structure, the query (\( Q \)) denotes the recipient of the message (i.e., the node or edge receiving the message), while the key (\( K \)) and value (\( V \)) are derived from the message sender (i.e., the node or edge transmitting the message). This design emphasizes the most pertinent message sources, enhancing the effectiveness of information dissemination across the graph. The attention mechanism is defined as follows:
\begin{equation}
  \text{Attention}(\mathbf{Q}, \mathbf{K}, \mathbf{V}) = \text{Softmax}\left(\frac{\mathbf{Q}\mathbf{K}^T}{\sqrt{d/K}}\right)\mathbf{V}
\end{equation}

Compared to existing MPNN structures, our HMG Encoder performs simultaneous Node Aggregation and Edge Aggregation, thereby synchronizing information propagation between nodes and edges more effectively.

\paragraph*{Node Aggregation}: **: For each node \( v_i \), our model gathers messages from connected edges \( e_{pi} \) using multi-head attention. This process aggregates message sources to update the representation of node \( v_i \):

\begin{equation}
  \begin{split}
    \mathbf{m}^{l}_{v_i} & = \text{AGG}^{(l)}(\{e^{l-1}_{pi}|{v_p} \in \mathcal{N}(v_i)\})                                                             \\
                         & = ( \|_{k=1}^{K} \sum \text{Attention}(\mathbf{q}^{l,k}_{v_i}, \mathbf{k}^{l,k}_{e_{pi}},\mathbf{v}^{l,k}_{e_{pi}})) W_{V};
  \end{split}
\end{equation}
Here, \( \|\ \) denotes concatenation in multi-head attention mechanisms, and \( W_{V} \in \mathbb{R}^{d \times d} \). \( \mathbf{m}^{l}_{v_i} \in \mathbb{R}^{d} \) represents the aggregated messages at node \( v_i \) during the \( l \)-th iteration of message passing. This aggregation incorporates messages from all incoming edges \( e_{pi} \) directed towards \( v_i \).

\paragraph*{Edge Aggregation}: Similarly, for each edge \( e_{ij} \), our model collects information from both the source node \( v_i \) and neighboring edges \( e_{pi} \). This mechanism captures relationships between nodes and edges, ensuring comprehensive edge representations:

\begin{equation}
  \begin{split}
    \mathbf{m}^{l}_{e_{ij}} & =  AGG^{(l)}(\psi \in \{v_i\}\cup \{e^{l-1}_{pi} : {v_p} \in \mathcal{N}(v_i)\})                                     \\
                            & = (\|_{k=1}^{K} \sum \text{Attention}(\mathbf{q}^{l,k}_{e_{ij}}, \mathbf{k}^{l,k}_\psi,\mathbf{v}^{l,k}_\psi)) W_{E}
  \end{split}
\end{equation}
Here, \( W_{E} \in \mathbb{R}^{d \times d} \). \( \mathbf{m}^{l}_{e_{ij}} \in \mathbb{R}^{d} \) denotes the messages aggregated for edge \( e_{ij} \), incorporating contributions from the source node \( v_i \) and all edges \( e_{pi} \) converging at \( v_i \). Importantly, the contribution from the reverse edge \( e_{ji} \) is inherently considered within the set \( \{ e_{pi} \} \), ensuring a comprehensive collection of edge-based information flows.

In each iteration, our model updates both nodes and edges simultaneously, ensuring the dynamic evolution of the global molecular representation. This approach enables continuous refinement of the graph understanding, capturing both micro-level interactions (e.g., atomic bonds) and macro-level properties (e.g., pharmacological attributes). This iterative process facilitates the model's ability to adapt and enhance its representation of complex molecular structures comprehensively.

\subsubsection*{Update Function}
Following the message-passing step, the update function integrates the incoming message vectors with the previous node or edge embeddings using a Multi-Layer Perceptron (MLP). This process ensures the seamless incorporation of new information into the existing representation, enabling the model to learn complex patterns while mitigating the risk of gradient vanishing or exploding:
\begin{equation}
  \begin{split}
    \mathbf{h}^{l}_{v_{i}} & = \text{UPDATE}(\mathbf{h}^{l-1}_{v_i}, \mathbf{m}^{l}_{v_i})                       \\
                           & = \text{LeakyReLU}((\mathbf{h}^{l-1}_{v_i} \| \mathbf{m}^{l}_{v_i})W^l_{\delta(v)})
  \end{split}
\end{equation}
Here, \( W^l \in \mathbb{R}^{2d \times d} \) denotes the update matrix, and the LeakyReLU activation function ensures consistency in the update process across nodes and edges, while accommodating the distinct attributes of different types.

\subsubsection*{Graph Readout}
In the final stage, we aggregate the learned node and edge representations into a unified graph representation using Self-Attention Graph Pooling (SAGPooling):

\begin{equation}
  \begin{split}
    \mathbf{h}_{\mathcal{G}} & = \text{READOUT}(\{\mathbf{h}^{L}_{v_{i}}\},(\{\mathbf{h}^{L}_{e_{ij}}\})) \\
                             & = \text{SAGPooling}(\mathbf{H}_{\mathcal{V}},\mathbf{H}_{\mathcal{E}})
  \end{split}
\end{equation}
This final aggregation step ensures that both node-level and edge-level information is preserved in the global graph representation, leading to a more comprehensive understanding of molecular properties.

\subsection*{Cross-view Contrastive Objective}

\subsubsection*{Batch Generation Strategy}

To utilize data from approved drug molecules effectively, we have devised a method to generate batch data for incorporation into our training procedures. This methodology guarantees that each batch is both balanced and representative, essential for training reliable predictive models in drug discovery and cheminformatics. Algorithm \ref{alg2} outlines the approach employed for generating training batch data.

\begin{algorithm}[h]
  \caption{Batch Generation}
  \label{alg2}
  \begin{algorithmic}[1]
    \REQUIRE The set $\mathcal{S}^M$ of molecules without drug ID and the set $\mathcal{S}^D$ of molecules with drug ID, Batch Size $N$, Complement Size $n$.
    \ENSURE Mini-Batches of size $N$.
    \STATE $\mathcal{S}^{U} = \{\}$
    \STATE $\text{Clusters},\text{Centers} = \text{KMeansConstrained}(\mathcal{S}^M , 0.7N)$
    \FOR {\( i = 0 \) to \ \text{Centers}.size}
    \STATE $\mathcal{D} = \text{NearestNeighbors}(\text{Centers}[i],\mathcal{S}^D, 0.3N - n)$
    \STATE $\text{Clusters}[i] \leftarrow \mathcal{D}$, $\mathcal{S}^{U} \leftarrow \mathcal{D}$
    \ENDFOR
    \FOR {\( i = 0 \) to \ \text{Centers}.size}
    \STATE $\text{Clusters}[i] \leftarrow \text{RandomSample}(\mathcal{S}^D- \mathcal{S}^U, n)$
    \ENDFOR
    \STATE \RETURN $\text{Clusters}$
  \end{algorithmic}
\end{algorithm}

The pre-training dataset was segregated into two subsets: one containing molecules with drug identifiers and another without. Molecules lacking drug IDs were clustered using a constrained K-means algorithm, with each cluster sized approximately at 70\% of the total dataset size \( N \). Subsequently, nearest neighbor searches were employed to pair each cluster with similar molecules possessing drug IDs, thereby adjusting the batch size to \( N - n \). Random molecules with drug IDs were then added to complete the batches. This method ensures that structurally similar molecules are grouped together, enhances the model's ability to discern molecular structures, mitigates sampling bias, and efficiently utilizes sparse drug data.

\subsubsection*{Cross-view Contrastive Loss}

Based on our batch generation strategy, for a mini-batch of size \( N \), we generate three sets of views as follows:
\begin{itemize}
  \item  \( \mathcal{S}^{M} = \{\mathcal{G}^{M}_1, \dots, \mathcal{G}^{M}_N \} \): This set includes the Molecule Views for all \( N \) molecules in the mini-batch.
  \item \( \mathcal{S}^{EM} = \{\mathcal{G}^{EM}_1, \dots, \mathcal{G}^{EM}_N \} \): This set comprises the Element Views for all \( N \) molecules in the mini-batch.
  \item \( \mathcal{S}^{DM} = \{\mathcal{G}^{DM}_1, \dots, \mathcal{G}^{DM}_{0.3N}\} \): This set includes the Drug Views for the last \( 0.3N \) molecules in the mini-batch that have Drug IDs.
\end{itemize}

The loss function \( \mathcal{L}_{(\mathcal{G}^{1},\mathcal{G}^{2},i)} \) for molecule \( i \), focused on views \( \mathcal{G}^{1} \) and \( \mathcal{G}^{2} \), is formulated as:
\begin{equation}\label{loss_function}
  \mathcal{L}_{(\mathcal{G}^{1},\mathcal{G}^{2},i)} = - \log \frac{e^{\text{sim}(\mathbf{z}_{\mathcal{G}^{1}_i}, \mathbf{z}_{\mathcal{G}^{2}_{\text{pos}}}) /\tau}}
  {\sum \iota e^{\text{sim}(\mathbf{z}_{\mathcal{G}^{1}_{i}}, \mathbf{z}_{\mathcal{G}^{1}_{\text{neg}}})/ \tau} +\sum  \iota e^{\text{sim}(\mathbf{z}_{\mathcal{G}^{1}_i}, \mathbf{z}_{\mathcal{G}^{2}_{\text{neg}}}) / \tau}}
\end{equation}
where \( \mathbf{z}_{\mathcal{G}^{1}_i} \) and \( \mathbf{z}_{\mathcal{G}^{2}_{\text{pos}}} \) are the embeddings of molecule \( i \) in \( \mathcal{G}^{1} \) and its positive pair in \( \mathcal{G}^{2} \). \( \mathbf{z}_{\mathcal{G}^{1}_{\text{neg}}} \) and \( \mathbf{z}_{\mathcal{G}^{2}_{\text{neg}}} \) are embeddings of negative samples for molecule \( i \) in \( \mathcal{G}^{1} \) and \( \mathcal{G}^{2} \). \( \text{sim}(\cdot, \cdot) \) denotes a similarity function measuring the similarity between embeddings. \( \tau \) is a temperature parameter that scales the logits for better convergence. The vector \( \mathbf{z}_{\mathcal{G}} \) is derived from \( \mathbf{h}_{\mathcal{G}} \) through the Projector head, expressed as:
\begin{equation}
  \mathbf{z}_{\mathcal{G}} = f_{\text{projector }}(\mathbf{h}_{\mathcal{G}})
\end{equation}

In the loss function provided, the terms and their roles are clarified as follows:
\begin{itemize}
  \item Positive Sample (Inter-Positive Contrastive) \( e^{\text{sim}(\mathbf{z}_{\mathcal{G}^{1}_i}, \mathbf{z}_{\mathcal{G}^{2}_{\text{pos}}}) / \tau } \): This term measures the similarity between the embedding \( \mathbf{z}_{\mathcal{G}^{1}_i} \) from \( \mathcal{G}^{1} \) and its corresponding positive pair \( \mathbf{z}_{\mathcal{G}^{2}_{\text{pos}}} \) from \( \mathcal{G}^{2} \).
  \item First Term in the Denominator (Intra-Negative Contrastive) \( e^{\text{sim}(\mathbf{z}_{\mathcal{G}^{1}_i}, \mathbf{z}_{\mathcal{G}^{1}_{\text{neg}}}) / \tau } \): This term sums over negative samples \( \mathbf{z}_{\mathcal{G}^{1}_{\text{neg}}} \) within \( \mathcal{G}^{1} \).
  \item Second  Term in the Denominator (Inter-Negative Contrastive) \( e^{\text{sim}(\mathbf{z}_{\mathcal{G}^{1}_i}, \mathbf{z}_{\mathcal{G}^{2}_{\text{neg}}}) / \tau } \): This term sums over negative samples \( \mathbf{z}_{\mathcal{G}^{2}_{\text{neg}}} \) within \( \mathcal{G}^{2} \).
\end{itemize}

Due to the inherent sparsity of drug molecules, the size of \(\mathcal{S}^{DM}\) is invariably less than that of \(\mathcal{S}^{M}\) and \(\mathcal{S}^{EM}\). Let us consider a view \(\mathcal{G}^{1}\) encompassing \(M\) samples and a view \(\mathcal{G}^{2}\) comprising \(N\) samples, where \(M \leq N\). In this context, the number of inter-positive contrastive pairs between the two views is \(M\), the number of inter-negative contrastive pairs associated with any given positive pair amounts to \(M-1\), and the number of intra-negative contrastive pairs is \(N-1\). It is important to observe that \(\mathcal{L}_{(\mathcal{G}^{1},\mathcal{G}^{2},i)} \neq \mathcal{L}_{(\mathcal{G}^{2},\mathcal{G}^{1},i)}\). Consequently, the loss function between \(\mathcal{G}^{1}\) and \(\mathcal{G}^{2}\) can be expressed as:

\begin{equation}\label{loss_function_views}
  \mathcal{L}^{1,2} = \frac{1}{M}\sum_{i=1}^M\mathcal{L}_{(\mathcal{G}^{1},\mathcal{G}^{2},i)} + \frac{1}{M}\sum_{i=1}^M\mathcal{L}_{(\mathcal{G}^{2},\mathcal{G}^{1},i)}
\end{equation}

\begin{figure}[!ht]
  \centerline{\includegraphics[width=0.65\textwidth]{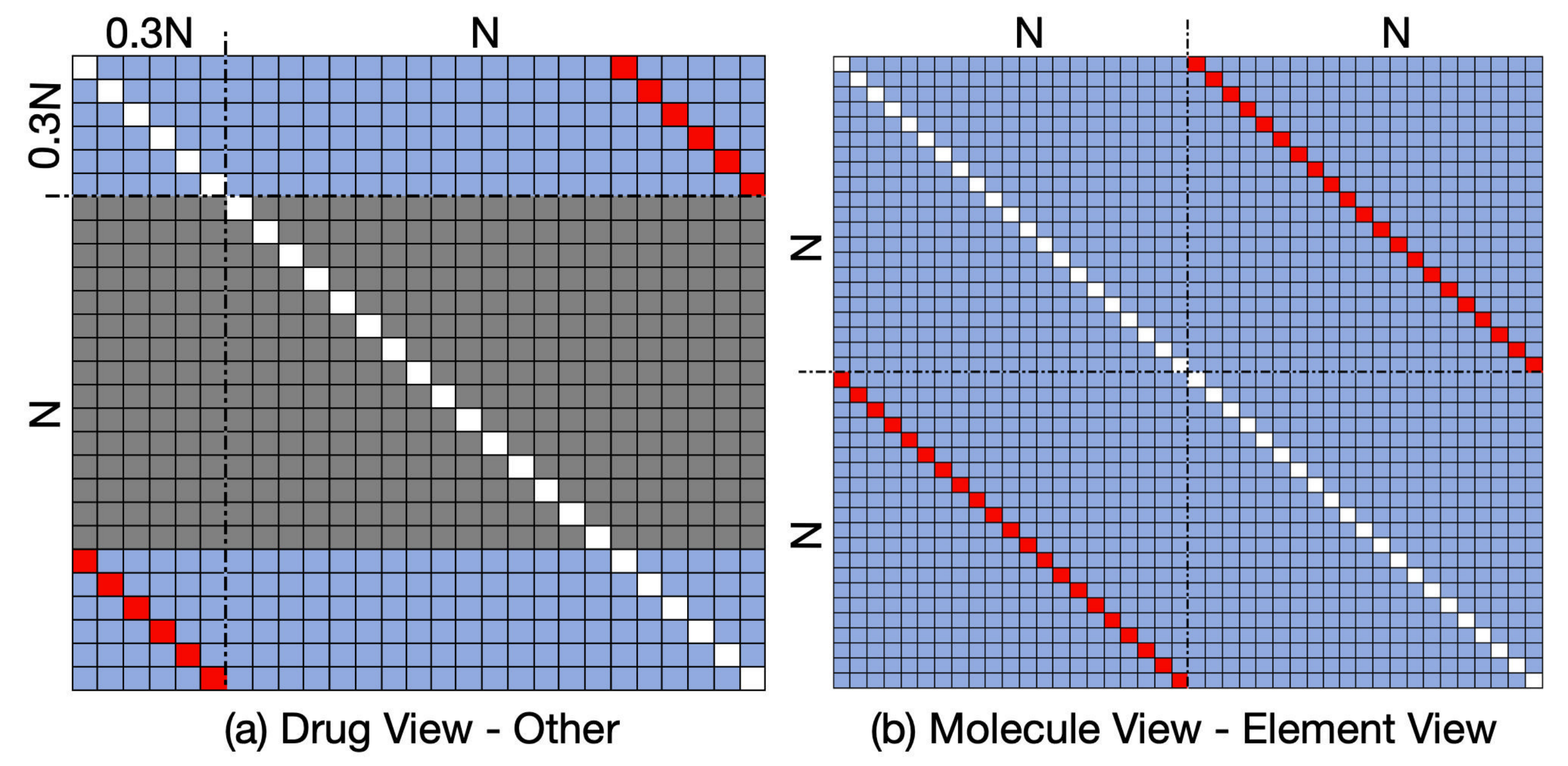}}
  \caption{Positive pairs and negative pairs between any two views. Red points for positive pairs and blue points for negative pairs. Each line forms a term in the loss function.}
  \label{cons}
\end{figure}

When either \(\mathcal{G}^{1}\) or \(\mathcal{G}^{2}\) is the drug view, the distribution of positive and negative samples within the loss function is depicted in Fig.\ref{cons}(a). In this figure, each row illustrates a sampled positive pair alongside its corresponding \(N+M-2\) negative sample pairs. Conversely, when neither \(\mathcal{G}^{1}\) nor \(\mathcal{G}^{2}\) is the drug view, the configuration of positive and negative samples is presented in Fig.\ref{cons}(b). In this scenario, the arrangement adheres to conventional practices in contrastive learning.

Thus, the comprehensive loss function encompassing all three views is expressed as:
\begin{equation}
  \mathcal{L}_\text{total} = \mathcal{L}^{M,EM} + \mathcal{L}^{M,DM} + \mathcal{L}^{EM,DM},
\end{equation}
where \(\mathcal{L}^{M,EM}\), \(\mathcal{L}^{M,DM}\), and \(\mathcal{L}^{EM,DM}\) denote the losses associated with the molecular–element, molecular–drug, and element–drug view pairings, respectively.

\subsection*{Fine-tuning on Molecular Property and DDI Prediction Tasks}

\begin{figure}[!ht] \centerline{\includegraphics[width=0.99\textwidth]{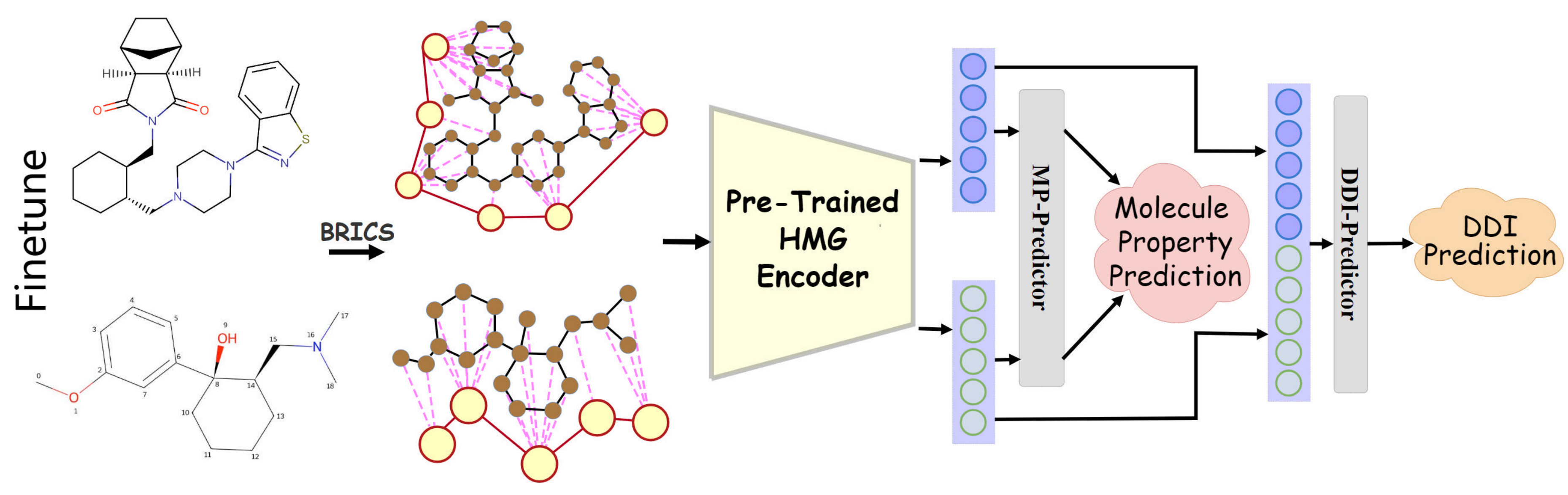}} \caption{Fine-tuning process for molecular property and DDI prediction tasks. The Projector used during pre-training is discarded, and the MP-Predictor and DDI-Predictor are employed for molecular property prediction and DDI prediction tasks, respectively.} \label{finetune} \end{figure}

The pre-trained molecular encoder is adaptable to a wide range of downstream molecular prediction tasks. As depicted in Fig.\ref{finetune}, the fine-tuning process demonstrates the application of the HMG Encoder for predicting both molecular properties and DDIs.

\subsubsection*{Molecular Property Prediction}
In the molecular property prediction task, every molecular graph \( \mathcal{G}\) is encoded by a HMG encoder. The overall graph representation \( \mathbf{h}_\mathcal{G}\) is then passed through a nonlinear predictor composed of fully connected layers, mapping the graph embedding to the target property:

\begin{equation}
  \hat{y}_{mp} = f_{\text{predictor}}(\mathbf{h}_{\mathcal{G}})
\end{equation}

\paragraph{DDI Prediction}
In the DDI  Prediction task, a pair of molecular graphs \( \mathcal{G}_1\) and\( \mathcal{G}_2\) are independently encoded using the same HMG encoder, resulting in the graph embeddings \( \mathbf{h}_{\mathcal{G}_1}\)and \( \mathbf{h}_{\mathcal{G}_2}\). This pair of embeddings are then concatenated into a joint representation:
\begin{equation}
  \mathbf{h}_{\text{pair}} = [\mathbf{h}_{\mathcal{G}_1}, \mathbf{h}_{\mathcal{G}_2}]
\end{equation}

This concatenated embedding is fed into a nonlinear predictor, which outputs the prediction of the interaction between the two drugs:

\begin{equation}
  \hat{y}_{ddi} = f_{\text{predictor}}(\mathbf{h}_{\text{pair}})
\end{equation}

\section*{Results and Discussion}

\subsection*{Experimental Setup}
\subsubsection*{Pre-training Phase}
In the pre-training phase, the KCHML model was initially trained on a comprehensive dataset comprising 250,000 unlabeled molecules sourced from ZINC15 \cite{sterling2015zinc}, along with 8,358 organic molecules from DRKG \cite{drkg2020}. DRKG is a comprehensive biological KG that connects genes, compounds, diseases, biological processes, side effects, and symptoms. It integrates information from six existing databases: DrugBank, Hestionet, GNBR, String, IntAct, and DGIdb.

\subsubsection*{Fine-tuning}
For the molecule property prediction tasks, we optimized computational efficiency by focusing solely on the molecule view encoder for downstream tasks, based on the premise that this encoder proficiently incorporates information from both the elemental and drug views. Additionally, we enhanced the SAGPooling graph readout layer by integrating newly designed SAGPooling and MLP layers specifically tailored to the tasks at hand.
We evaluated the model using 13 benchmark datasets from MoleculeNet \cite{wu2018moleculenet}, which encompass a wide array of molecular data across various scientific disciplines. To ensure robust and reliable performance, we adhered to standard practices by employing 5-fold cross-validation with an 8:1:1 train/validation/test split, and conducted three independent training runs.

For the DDI prediction task, given that the DRKG database already includes some data from DrugBank, we opted to exclude DrugBank from our experiments and instead evaluated the model performance solely on the TwoSide dataset. We conduct experiments on the Twoside dataset under two different settings: the transductive scenario and the inductive scenario. In the transductive setting, drugs appear in both the training and test sets, which allows for a direct evaluation of the model's performance on seen drugs. In contrast, the inductive setting includes drugs that are either entirely absent or only partially represented in the training set, which enables us to assess the model's ability to generalize to unseen drugs. In particular, in the inductive setting, we explore two distinct scenarios: Old-new drug pairs and new-new drug pairs, to further analyze the performance of each model. We follow the setting and negative sample generation strategy from GMPNN~\cite{10.1093/bib/bbab441}, a method specifically designed for DDI prediction.

\subsection*{Baselines}
\paragraph*{Supervised Learning Baselines}:
We benchmarked KCHML against several well-established graph neural network architectures, including GCN \cite{kipf2016semi}, GIN \cite{xu2018powerful}, and AttentiveFP \cite{xiong2019pushing}. Additionally, we compared it with two variants of message-passing neural networks (MPNNs) specifically tailored for molecular data: DMPNN \cite{yang2019analyzing} and CMPNN \cite{song2020communicative}. CoMPT \cite{ijcai2021p0309} was also included for its consideration of edge features and its enhancement of message interactions between bonds and atoms.

\paragraph*{Pre-trained Methods}:
In the realm of predictive-based self-supervised learning, we incorporated several pre-training models for comparison: N-GRAM \cite{liu2019n}, which constructs node embeddings through short walks and employs Random Forest or XGBoost for property prediction; Hu et al. \cite{hu2020strategies} and GROVER \cite{rong2020self}, which integrate both node-level and graph-level knowledge in their pretext tasks.

\paragraph*{Graph Contrastive Learning Baselines}:
We evaluated KCHML against existing graph contrastive learning frameworks. MoCL \cite{sun2021mocl} leverages domain knowledge at two distinct levels to enhance representation learning. MolCLR \cite{MolCLR} applies general graph augmentation techniques to molecular data. KCL \cite{fang2022molecular} utilizes a chemical element KG to augment the original molecular graph.

\paragraph*{DDI-specific Models}:
We evaluated KCHML against existing DDI-specific models on DDI prediction tasks. DeepDDI~\cite{ryu2018deep} and GMPNN~\cite{10.1093/bib/bbab441} focus on leveraging molecular graph representations to model the interactions between drug pairs, while SA-DDI~\cite{yang2022learning} uses attention mechanisms to enhance the model’s ability to focus on relevant drug features. DGNN-DDI~\cite{ma2023dual} incorporates dual graph neural networks to capture the rich relationship between drug molecules and their targets. TIGER~\cite{su2024dual}, on the other hand, integrates multi-modal data to better understand the interactions between drugs, genes, and diseases, providing a more holistic approach to DDI prediction.

\subsection*{Overall Performance}

In this section, we examine the KCHML model's overall performance on molecule property prediction and DDI prediction across transductive and inductive settings.

\subsubsection*{Molecule Property Prediction}

Table \ref{table_result1} and \ref{table_result2} display the performance of various models across 13 datasets for molecule property prediction. The following conclusions can be drawn regarding the performance of various models:

\begin{table*}[!ht]
  \begin{adjustwidth}{-1.5in}{0in}
    \centering
    \caption{Comparison of the models on classification datasets for molecule property prediction. The best-performing results are highlighted in bold.}
    \resizebox{1.25\textwidth}{!}{
      \label{table_result1}
      \begin{tabular}{l|cccccccc}
        \hline
        \textbf{Task}         & \multicolumn{8}{c}{\textbf{Classification (ROC-AUC)} $\uparrow$}                                                                                                                                                                        \\
        \hline
        \textbf{Dataset}      & \textbf{BBBP}                                                    & \textbf{Tox21}       & \textbf{SIDER}        & \textbf{ClinTox}      & \textbf{ToxCast}      & \textbf{BACE}         & \textbf{MUV}          & \textbf{HIV}          \\
        \textbf{\# Molecules} & 2,039                                                            & 7,831                & 1,427                 & 1,478                 & 8,575                 & 1,513                 & 93,087                & 41,127                \\
        \textbf{\# Tasks}     & 1                                                                & 12                   & 27                    & 2                     & 617                   & 1                     & 17                    & 1                     \\
        \hline
        GCN                   & 71.00 ± 0.91                                                     & 69.96 ± 0.30         & 54.07 ± 0.30          & 62.14 ± 2.80          & 64.32 ± 6.06          & 71.83 ± 3.97          & 72.12 ± 4.01          & 74.02 ± 2.97          \\

        GIN                   & 65.65 ± 4.45                                                     & 74.17 ± 0.79         & 56.87 ± 1.58          & 57.06 ± 4.41          & 65.99 ± 1.48          & 71.49 ± 2.49          & 74.89 ± 1.88          & 75.62 ± 1.87          \\
        AttentiveFP           & 89.11 ± 0.50                                                     & 79.97 ± 2.04         & 60.18 ± 0.59          & 92.35 ± 2.35          & 58.36 ± 0.10          & 86.11 ± 1.50          & 78.29 ± 3.84          & 75.88 ± 1.38          \\
        MPNN                  & 91.85 ± 2.56                                                     & 82.34 ± 2.58         & 57.03 ± 2.02          & 86.18 ± 1.10          & 66.33 ± 2.26          & 82.04 ± 3.93          & 77.91 ± 4.22          & 78.59 ± 3.96          \\
        DMPNN                 & 85.40 ± 3.53                                                     & 82.67 ± 2.20         & 58.55 ± 1.69          & 86.03 ± 2.89          & 66.20 ± 2.07          & 83.39 ± 2.79          & 79.30 ± 2.76          & 80.73 ± 1.72          \\
        CMPNN                 & 90.51 ± 3.93                                                     & 81.32 ± 2.72         & 64.59 ± 0.79          & 88.95 ± 2.10          & 68.40 ± 0.59          & 91.69 ± 3.14          & 80.77 ± 2.42          & 80.78 ± 1.60          \\
        CoMPT                 & 94.57 ± 1.20                                                     & 80.91 ± 1.47         & 63.86 ± 2.81          & 90.20 ± 1.92          & 66.42 ± 2.11          & 82.47 ± 0.69          & 80.29 ± 4.52          & 78.63 ± 2.61          \\
        \hline
        N-GRAM                & 91.03 ± 0.30                                                     & 77.41 ± 2.72         & -                     & 88.27 ± 2.69          & -                     & 78.04 ± 1.28          & 75.44 ± 0.70          & 77.78 ± 0.39          \\
        Hu et al.             & 70.85 ± 1.50                                                     & 77.50 ± 0.40         & 62.74 ± 0.80          & 72.38 ± 1.51          & 65.25 ± 0.59          & 84.26 ± 0.69          & 81.83 ± 2.06          & 79.76 ± 0.70          \\
        GROVER                & 84.37 ± 4.10                                                     & 80.79 ± 1.97         & 57.06 ± 1.51          & 71.02 ± 7.24          & 56.07 ± 0.50          & 82.34 ± 8.83          & 69.40 ± 1.48          & 67.80 ± 1.49          \\ \hline
        MolCL                 & 87.92 ± 1.78                                                     & 76.98 ± 1.51         & 61.57 ± 4.20          & 79.42 ± 2.08          & 64.86 ± 1.86          & 84.34 ± 0.78          & 79.58 ± 2.28          & 76.66 ± 0.60          \\
        MolCLR                & 72.62 ± 1.00                                                     & 74.38 ± 5.32         & 61.37 ± 3.63          & 90.41 ± 2.66          & 65.06 ± 2.10          & 82.30 ± 0.71          & 81.27 ± 4.58          & 78.68 ± 0.59          \\
        KCL                   & 95.38 ± 1.70                                                     & 85.23 ± 5.27         & 65.84 ± 3.62          & 94.98 ± 2.65          & 75.19 ± 2.09          & 93.00 ± 0.69          & 82.88 ± 2.30          & 84.80 ± 0.60          \\ \hline
        KCHML                 & \textbf{95.89 ± 1.66}                                            & \textbf{85.83± 3.41} & \textbf{69.41 ± 0.96} & \textbf{96.12 ± 2.14} & \textbf{75.97 ± 1.25} & \textbf{94.57 ± 1.63} & \textbf{83.14 ± 1.89} & \textbf{85.19 ± 0.85} \\ \hline
      \end{tabular}
    }
  \end{adjustwidth}
\end{table*}

\begin{table*}[!ht]
  \begin{adjustwidth}{-1in}{0in}
    \centering
    \caption{Comparison of the models on regression datasets for molecule property prediction. The best-performing results are highlighted in bold.}
    \label{table_result2}
    \begin{tabular}{l|ccc|ccc}
      \hline
      \textbf{Task}         & \multicolumn{3}{c|}{\textbf{Regression (RMSE)} $\downarrow$} & \multicolumn{2}{c}{\textbf{Regression (MSE) }$\downarrow$}                                                                          \\
      \hline
      \textbf{Dataset}      & \textbf{ESOL}                                                & \textbf{FreeSolv}                                          & \textbf{Lipo}          & \textbf{QM7}        & \textbf{QM8}            \\
      \textbf{\# Molecules} & 1,128                                                        & 642                                                        & 4,200                  & 6,830               & 21,786                  \\
      \textbf{\# Tasks}     & 1                                                            & 1                                                          & 1                      & 1                   & 12                      \\
      \hline
      GCN                   & 1.417 ± 0.050                                                & 2.887 ± 0.133                                              & 0.700 ± 0.049          & 123.3 ± 2.2         & 0.0365 ± 0.000          \\
      GIN                   & 1.440 ± 0.020                                                & 2.785 ± 0.177                                              & 0.854 ± 0.071          & 123.9 ± 0.7         & 0.0373 ± 0.001          \\
      AttentiveFP           & 2.089 ± 0.182                                                & 0.879 ± 0.029                                              & 0.714 ± 0.001          & 103.1 ± 0.9         & 0.0184 ± 0.001          \\
      MPNN                  & 1.155 ± 0.428                                                & 1.611 ± 0.957                                              & 0.665 ± 0.051          & 111.4 ± 0.9         & 0.0148 ± 0.001          \\
      DMPNN                 & 1.049 ± 0.008                                                & 1.654 ± 0.081                                              & 0.682 ± 0.016          & 103.5 ± 8.6         & 0.0153 ± 0.001          \\
      CMPNN                 & 0.783 ± 0.111                                                & 1.560 ± 0.439                                              & 0.610 ± 0.029          & 74.5 ± 3.1          & 0.0153 ± 0.002          \\
      CoMPT                 & 0.832 ± 0.039                                                & 1.940 ± 0.808                                              & 0.647 ± 0.028          & 86.5 ± 1.3          & 0.0187 ± 0.001          \\ \hline
      N-GRAM                & 1.107 ± 0.030                                                & 2.472 ± 0.192                                              & 0.887 ± 0.121          & 125.5 ± 1.5         & 0.0317 ± 0.003          \\
      Hu et al.             & 1.100 ± 0.006                                                & 2.714 ± 0.002                                              & 0.725 ± 0.003          & 113.6 ± 0.6         & 0.0212 ± 0.001          \\
      GROVER                & 1.435 ± 0.283                                                & 2.935 ± 0.620                                              & 0.829 ± 0.010          & 90.0 ± 1.9          & 0.0180 ± 0.001          \\ \hline
      MolCL                 & 1.038 ± 0.270                                                & 1.884 ± 0.266                                              & 0.662 ± 0.008          & 99.4 ± 3.7          & 0.0180 ± 0.001          \\
      MolCLR                & 1.105 ± 0.023                                                & 2.255 ± 0.246                                              & 0.779 ± 0.009          & 91.2 ± 1.7          & 0.0184 ± 0.013          \\
      KCL                   & 0.659 ± 0.019                                                & 1.148 ± 0.257                                              & 0.566 ± 0.007          & 59.9 ± 2.8          & 0.0130 ± 0.013          \\ \hline
      KCHML                 & \textbf{0.612 ± 0.142}                                       & \textbf{1.136± 0.142}                                      & \textbf{0.527 ± 0.009} & \textbf{56.1 ± 3.5} & \textbf{0.0121 ± 0.000} \\ \hline
    \end{tabular}
  \end{adjustwidth}
\end{table*}

\begin{itemize}
  \item  \textbf{KCHML's Superior Performance}: Across both classification (Table \ref{table_result1}) and regression tasks (Table \ref{table_result2}), the KCHML model consistently outperformed all others. It achieved the highest ROC-AUC scores for classification and the lowest RMSE for regression tasks, demonstrating its strong predictive capabilities for molecular properties. This highlights the effectiveness of KCHML’s architecture in integrating advanced learning algorithms with complex molecular data.
  \item \textbf{Leverage of KGs}: Models like KCHML and KCL, which incorporate knowledge graphs, showed a clear advantage over methods that do not. This is attributed to the rich semantic information provided by KGs, enabling these models to capture intricate molecular interactions and properties more effectively. The integration of structured data through KGs significantly enhanced their prediction accuracy.
  \item \textbf{Comparison with Contrastive Learning Models}: KCHML demonstrated marked improvements over models such as MoCL and MolCLR, both of which employ contrastive learning but do not utilize domain knowledge. Unlike MoCL’s augmentation strategies, which may disrupt molecular integrity and introduce noise, KCHML uses more refined augmentation techniques that preserve the structure and function of the molecules. This approach leads to more accurate and stable learning, reducing the risk of misleading training signals.
\end{itemize}

\subsubsection*{DDI Prediction in Transductive Setting}

In the DDI prediction task with transductive setting, we retained only the models that performed well in molecular property prediction and compared them against five baseline models specifically designed for DDI prediction: DeepDDI~\cite{ryu2018deep}, GMPNN~\cite{10.1093/bib/bbab441}, SA-DDI~\cite{yang2022learning}, DGNN-DDI~\cite{ma2023dual} and TIGER~\cite{su2024dual}. Table~\ref{table_result3_transductive} summarizes the performance of each model across various metrics on TwoSide dataset, from which we drew the following conclusions.

\begin{table*}[!ht]
  \centering
  \caption{Comparison of the DDI prediction models for the transductive setting on TwoSide dataset. The best-performing results are highlighted in bold.}
  \label{table_result3_transductive}
  \begin{tabular}{l|cccc}
    \hline
    \textbf{Metric} & \textbf{ACC}$\uparrow$ & \textbf{AUC}$\uparrow$ & \textbf{AP}$\uparrow$ & \textbf{F1}$\uparrow$ \\ \hline
    DMPNN           & 72.13 ± 0.20           & 76.85 ± 0.31           & 75.95 ± 0.16          & 72.64 ± 0.32          \\
    CMPNN           & 73.06 ± 0.38           & 77.90 ± 0.16           & 76.41 ± 0.22          & 73.48 ± 0.25          \\
    CoMPT           & 74.60 ± 0.36           & 79.50 ± 0.25           & 77.03 ± 0.16          & 77.23 ± 0.22          \\ \hline
    N-GRAM          & 73.85 ± 0.25           & 78.90 ± 0.30           & 75.72 ± 0.23          & 73.99 ± 0.39          \\
    Hu et al.       & 74.10 ± 0.31           & 80.05 ± 0.14           & 75.91 ± 0.26          & 74.64 ± 0.27          \\
    GROVER          & 74.50 ± 0.28           & 79.25 ± 0.19           & 76.06 ± 0.19          & 75.31 ± 0.19          \\ \hline
    MolCL           & 75.30 ± 0.23           & 80.10 ± 0.22           & 78.70 ± 0.21          & 75.15 ± 0.16          \\
    MolCLR          & 76.05 ± 0.31           & 79.15 ± 0.35           & 77.84 ± 0.32          & 77.44 ± 0.18          \\
    KCL             & 77.96 ± 0.22           & 86.97 ± 0.16           & 82.36 ± 0.23          & 80.41 ± 0.25          \\ \hline
    DeepDDI         & 70.52 ± 0.27           & 76.96 ± 0.19           & 75.18 ± 0.31          & 74.08 ± 0.15          \\
    GMPNN           & 82.83 ± 0.14           & 90.07 ± 0.12           & 87.24 ± 0.11          & 84.08 ± 0.23          \\
    SA-DDI          & 82.89 ± 0.10           & 90.75 ± 0.13           & 88.98 ± 0.18          & 84.11 ± 0.20          \\
    DGNN-DDI        & 83.32 ± 0.12           & \textbf{91.28 ± 0.14}  & 88.58 ± 0.12          & \textbf{85.18 ± 0.18} \\
    TIGER           & \textbf{83.77 ± 0.11}  & 90.80 ± 0.14           & 89.17 ± 0.15          & 84.34 ± 0.22          \\
    \hline
    KCHML           & 81.23 ± 0.11           & 90.21 ± 0.14           & \textbf{89.21 ± 0.25} & 82.89 ± 0.31          \\ \hline
  \end{tabular}
\end{table*}

\begin{itemize}
  \item Generally, DDI-specific models outperform those designed for molecular property prediction: When comparing models specifically designed for DDI prediction (e.g., GMPNN, SA-DDI, DGNN-DDI, and TIGER) to those originally intended for molecular property prediction (e.g., DMPNN, CMPNN, MolCL, and KCL), we observe that the DDI-specific models tend to perform better across most metrics, particularly in accuracy, AUC, and F1 score. This is expected since these models have been fine-tuned for the nuances of DDI prediction, which requires specialized understanding of drug interactions, as opposed to general molecular property features.
  \item KCHML outperforms all models designed for molecular property prediction, including pre-trained models: Among the models that were originally designed for molecular property prediction, KCHML demonstrates the strongest performance, particularly excelling in average precision (AP), where it achieves the highest score of 89.21 ± 0.25. This shows that KCHML is not only competitive with models developed specifically for molecular properties but also surpasses many in key metrics. Despite not being optimized for DDI prediction, KCHML’s performance indicates its robustness in identifying relevant interactions, making it highly effective for DDI prediction tasks.
  \item KCHML remains highly competitive when compared to DDI-specific models: Even when compared to models designed specifically for DDI prediction, KCHML shows a remarkable level of competitiveness. Although DGNN-DDI and TIGER perform better in terms of accuracy and F1 score, KCHML still holds strong, particularly in average precision, where it achieves the best result. The fact that KCHML maintains high performance in terms of precision suggests it is a solid contender for practical DDI prediction tasks, where the goal is to minimize false positives and maximize the identification of relevant drug interactions.
\end{itemize}

\subsubsection*{DDI Prediction in Inductive Setting}

\begin{table*}[!ht]
  \begin{adjustwidth}{-1.2in}{0in}
    \caption{Comparison of the DDI prediction models for the inductive setting on TwoSide dataset. The best-performing results are highlighted in bold.}
    \resizebox{1.25\textwidth}{!}{
      \label{table_result3_inductive}
      \begin{tabular}{l|llll|llll}
        \hline
        \textbf{Partition} & \multicolumn{4}{c|}{\textbf{old-new}} & \multicolumn{4}{c}{\textbf{new-new}}                                                                                                                                                 \\ \hline
        \textbf{Metric}    & \textbf{ACC }                         & \textbf{AUC }                        & \textbf{AP }          & \textbf{F1 }          & \textbf{ACC }         & \textbf{AUC }         & \textbf{AP }          & \textbf{F1 }          \\
        DMPNN              & 68.69 ± 0.59                          & 72.9 ± 0.39                          & 72.14 ± 1.15          & 69.29 ± 0.18          & 63.99 ± 0.70          & 69.95 ± 0.69          & 69.52 ± 0.73          & 63.71 ± 0.51          \\
        CMPNN              & 70.08 ± 1.25                          & 73.61 ± 0.77                         & 74.51 ± 0.65          & 69.88 ± 0.60          & 64.44 ± 0.31          & 70.31 ± 0.49          & 69.97 ± 0.81          & 62.92 ± 0.33          \\
        CoMPT              & 73.66 ± 0.47                          & 76.40 ± 0.31                         & 75.06 ± 1.30          & 72.48 ± 0.57          & 68.76 ± 0.82          & 74.67 ± 0.24          & 73.83 ± 1.29          & 67.89 ± 0.98          \\ \hline
        N-GRAM             & 74.29 ± 0.78                          & 76.81 ± 0.60                         & 75.75 ± 0.77          & 71.94 ± 0.92          & 70.07 ± 1.00          & 75.53 ± 0.49          & 73.93 ± 0.74          & 66.94 ± 0.28          \\
        Hu et al.          & 74.29 ± 0.55                          & 78.20 ± 0.81                         & 75.89 ± 0.95          & 72.29 ± 0.22          & 68.67 ± 0.22          & 76.10 ± 1.16          & 74.83 ± 0.75          & 70.74 ± 1.11          \\
        GROVER             & 73.97 ± 0.47                          & 77.62 ± 0.33                         & 76.21 ± 0.23          & 73.14 ± 0.43          & 68.30 ± 1.03          & 75.35 ± 0.89          & 74.72 ± 1.31          & 69.89 ± 0.17          \\ \hline
        MolCL              & 74.61 ± 0.82                          & 77.93 ± 0.26                         & 76.45 ± 0.22          & 73.17 ± 0.97          & 69.64 ± 1.12          & 74.75 ± 0.49          & 73.96 ± 0.21          & 68.31 ± 0.72          \\
        MolCLR             & 74.61 ± 1.19                          & 77.72 ± 0.66                         & 75.71 ± 0.23          & 73.03 ± 0.38          & 70.27 ± 1.16          & 74.10 ± 0.57          & 73.43 ± 1.20          & 67.71 ± 1.26          \\
        KCL                & 76.31 ± 0.86                          & \textbf{84.27 ± 0.61}                & 80.71 ± 1.18          & 77.61 ± 1.16          & 71.96 ± 0.77          & 81.60 ± 0.96          & 80.04 ± 0.80          & 71.76 ± 1.23          \\ \hline
        DeepDDI            & 63.79 ± 1.07                          & 71.97 ± 1.00                         & 70.63 ± 0.93          & 66.70 ± 0.65          & 58.97 ± 0.62          & 67.42 ± 0.20          & 66.29 ± 0.26          & 61.38 ± 0.89          \\
        GMPNN              & 74.55 ± 0.92                          & 80.91 ± 0.80                         & 78.58 ± 0.59          & 73.93 ± 0.86          & 68.49 ± 0.33          & 77.69 ± 0.26          & 76.52 ± 0.76          & 71.74 ± 0.28          \\
        DGNN-DDI           & 74.57 ± 0.34                          & 81.96 ± 0.26                         & 79.55 ± 0.27          & 75.73 ± 0.19          & 70.81 ± 0.33          & 78.61 ± 0.30          & 77.24 ± 0.17          & 73.33 ± 0.76          \\
        DSN-DDI            & 74.54 ± 0.55                          & 81.59 ± 0.66                         & 78.86 ± 0.24          & 74.84 ± 0.43          & 69.85 ± 0.54          & 77.25 ± 0.23          & 75.96 ± 0.24          & 72.99 ± 0.64          \\
        TIGER              & /                                     & /                                    & /                     & /                     & /                     & /                     & /                     & /                     \\  \hline
        KCHML              & \textbf{77.11 ± 0.62}                 & 83.75 ± 0.89                         & \textbf{81.21 ± 0.24} & \textbf{79.23 ± 0.30} & \textbf{72.69 ± 0.51} & \textbf{81.87 ± 0.54} & \textbf{79.86 ± 0.65} & \textbf{74.27 ± 0.42} \\ \hline
      \end{tabular}
    }
  \end{adjustwidth}
\end{table*}

In the inductive Setting, the task involves predicting DDIs for unseen drug pairs, which presents a more challenging scenario compared to the transductive setting. The inductive setting tests the model’s ability to generalize to novel drug combinations that were not part of the training set. This setting is more aligned with real-world scenarios where new drug pairs are constantly being discovered, and the model must predict interactions without prior exposure to those pairs. Since the TIGER model cannot handle the cold-start problem, we did not include it in the inductive setting.  Based on the analysis of the data presented in Table~\ref{table_result3_inductive}, the following conclusions can be drawn:
\begin{itemize}
  \item All Models Experience Performance Decline in the Inductive Setting Compared to the Transductive Setting, with a Further Drop in the New-New Partition: Across all models, there is a noticeable decline in performance when comparing the inductive setting to the transductive setting. This is observed in both accuracy and AUC metrics, where models generally perform worse in the inductive setting due to the challenge of generalizing to unseen drug pairs. Additionally, the new-new partition sees a further drop in performance compared to the old-new partition, indicating that models struggle more when predicting interactions between entirely new drugs, further emphasizing the difficulty of the inductive task.
  \item DDI-Specific Models Experience a Greater Decline Compared to Molecular Property Prediction Models: In the inductive setting, DDI-specific models (e.g., KCL, GMPNN, DGNN-DDI) experience a more significant drop in performance compared to molecular property prediction models (e.g., DMPNN, CMPNN, MolCL). While DDI-specific models perform better than molecular property prediction models in the transductive setting, this advantage becomes less pronounced in the inductive setting, where both types of models show comparable performance. The decline in performance is more substantial for the DDI-specific models, which may suggest that their reliance on specific training data related to drug interactions is less effective in the inductive setting where the data distribution changes.
  \item KCHML Exhibits Clear Performance Advantages in the Inductive Setting, Performing Outstandingly in Old-New and Best in New-New: Among the molecular property prediction models, KCHML shows a distinct performance advantage, especially in the inductive setting. In the old-new partition, KCHML already performs excellently, achieving the highest accuracy (77.11 ± 0.62) and F1 score (79.23 ± 0.30). In the new-new partition, KCHML further stands out, achieving the best performance across all metrics, including accuracy (72.69 ± 0.51), AUC (81.87 ± 0.54), average precision (79.86 ± 0.65), and F1 score (74.27 ± 0.42). This demonstrates that KCHML not only excels among molecular property prediction models but also competes strongly in the inductive setting, particularly when predicting interactions between entirely new drug pairs.
\end{itemize}
In summary, KCHML is likely more adaptable to unseen data in the inductive setting. This adaptability is crucial when the model encounters new drug pairs that have not been seen during training. Unlike DDI-specific models, KCHML has been trained on a broader range of molecular property tasks, enabling it to generalize better when faced with novel drug pairs.
Additionally, KCHML benefits from being trained within a multi-task learning framework that incorporates various molecular property prediction tasks. This framework provides KCHML with a richer set of learned features, which helps it capture a broader range of patterns that can be applied to DDI prediction, even when faced with previously unseen drug pairs. The generalizability learned across multiple tasks enables KCHML to perform more effectively in the inductive setting, where the data distribution may be different from the training data.

\subsection*{Ablation Experiments}
To explore the impact of different encoders, we replaced our HMG encoder with other encoders such as RGCN, MPNN, DMPNN, and CMPNN to form $\text{KCHML}_{R}$, $\text{KCHML}_{M}$, $\text{KCHML}_{D}$, and $\text{KCHML}_{C}$, respectively.
Table \ref{table_result_encoder} shows the performance values on the molecule property
prediction task.
\begin{table}[!ht]
  \centering
  \caption{Comparison of different encoders in molecule property
    prediction tasks. The best results are highlighted in bold and the suboptimal results are marked with *.}
  \label{table_result_encoder}
  \begin{tabular}{l|cccccccc}
    \hline
    \textbf{Model}     & \textbf{BBBP}  & \textbf{Tox21} & \textbf{SIDER}     & \textbf{ClinTox}   & \textbf{ToxCast}   & \textbf{BACE}      & \textbf{MUV}       & \textbf{HIV}       \\
    \hline
    $\text{KCHML}_{R}$ & 92.11          & 82.36          & 67.26              & 92.69              & 72.66              & 90.93              & 79.74              & 82.96              \\
    $\text{KCHML}_{M}$ & 94.43          & 82.8           & $\text{68.42}^{*}$ & 94.05              & 74.37              & 92.84              & 82.65              & 83.79              \\
    $\text{KCHML}_{D}$ & 92.56          & 82.67          & 67.23              & 94.44              & 73.89              & 93.06              & 81.75              & 82.34              \\
    $\text{KCHML}_{C}$ & 95.17          & 85.18          & 67.79              & $\text{95.78}^{*}$ & $\text{75.56}^{*}$ & $\text{94.38}^{*}$ & $\text{82.94}^{*}$ & $\text{85.01}^{*}$ \\ \hline
    $\text{KCHML}$     & \textbf{95.89} & \textbf{85.83} & \textbf{69.41}     & \textbf{96.12}     & \textbf{75.97}     & \textbf{94.57}     & \textbf{83.14}     & \textbf{85.19}     \\
    \hline
  \end{tabular}
\end{table}

The following conclusions can be observed:
\begin{itemize}
  \item HMG Encoder Outperforms Other Encoders: The HMG encoder within the KCHML framework consistently yields the best performance across all datasets, as indicated by the bolded results in the table. Specifically, KCHML with the HMG encoder achieves the highest performance in terms of accuracy across datasets such as BBBP, Tox21, SIDER, ClinTox, ToxCast, BACE, MUV, and HIV. The superior results suggest that the HMG encoder is particularly effective at learning rich and informative representations, enhancing the model's ability to predict drug interactions and toxicity.
  \item Benefit of the Multi-View Contrastive Learning Framework: A deeper comparison of the results presented in Table \ref{table_result1} and Table \ref{table_result_encoder} highlights a significant improvement in performance for all encoders (including MPNN, DMPNN, and CMPNN) using multi-view contrastive learning framework. This indicates that the multi-view contrastive learning framework plays a crucial role in enhancing the performance of these encoders. By incorporating contrastive learning, the models are able to capture more comprehensive molecular representations, contributing to better predictive performance across all datasets.
\end{itemize}

The impact of removing individual views, namely the element view and the drug view, from the KCHML model was also evaluated. The modified models were denoted as $\text{KCHML}_{w/oE}$ and $\text{KCHML}_{w/oD}$, corresponding to the absence of the element view and the drug view, respectively. The results are summarized in Table \ref{table_result_view}, which shows that the model's performance is significantly affected when the element view is removed, and less so when the drug view is omitted. Below is a detailed analysis of these results.

\begin{table}[!ht]
  \centering
  \caption{Comparison of different views in molecule property
    prediction tasks. The best results are marked in bold.}
  \label{table_result_view}
  \begin{tabular}{l|cccccccc}
    \hline
    \textbf{Model}        & \textbf{BBBP}  & \textbf{Tox21} & \textbf{SIDER} & \textbf{ClinTox} & \textbf{ToxCast} & \textbf{BACE}  & \textbf{MUV}   & \textbf{HIV}   \\
    \hline
    $\text{KCHML}_{w/oE}$ & 89.71          & 79.92          & 64.52          & 93.29            & 72.32            & 90.7           & 81.78          & 80.3           \\
    $\text{KCHML}_{w/oD}$ & 94.96          & 85.53          & 66.94          & 95.91            & 75.77            & 93.97          & 82.94          & 84.89          \\ \hline
    $\text{KCHML}$        & \textbf{95.89} & \textbf{85.83} & \textbf{69.41} & \textbf{96.12}   & \textbf{75.97}   & \textbf{94.57} & \textbf{83.14} & \textbf{85.19} \\
    \hline
  \end{tabular}
\end{table}

\begin{itemize}
  \item Removing the Element View ($\text{KCHML}_{w/oE}$) Leads to a Significant Performance Drop: When the element view is removed, there is a marked decrease in performance across nearly all datasets. Specifically, $\text{KCHML}_{w/oE}$ exhibits a noticeable reduction in accuracy in comparison to the full KCHML model (denoted as KCHML), especially in BBBP, SIDER, and HIV, where the performance is reduced by several percentage points. This significant decline suggests that the element view is a crucial component of the model for capturing fundamental chemical properties and interactions within the molecular structure. The element view likely provides detailed information about atomic-level interactions, bond types, and molecular fragments, which are essential for understanding how molecules interact at a fundamental level. Removing this view hampers the model's ability to learn these intricate molecular details, leading to a substantial loss in predictive accuracy.
  \item Removing the Drug View ($\text{KCHML}_{w/oD}$) Affects Performance but Less Severely: On the other hand, removing the drug view results in a performance decline, but it is less severe compared to the loss of the element view. While the drug view is important for providing high-level insights into the overall drug interactions and functional properties, its absence does not completely negate the model's effectiveness. The drug view likely encodes broader characteristics such as drug classes, drug targets, and their functional role in biological systems. However, the performance decline here is smaller than that observed in the absence of the element view, indicating that the drug view, while valuable, is somewhat secondary to the element view for the task at hand.
  \item Root Cause of Poor Performance Without the Element View: On one hand, the element view provides the model with sufficient atomic-level details (including atomic interactions and molecular substructures) that are critical for maintaining strong predictive capabilities. These details are essential for predicting molecular properties such as toxicity, binding affinity, and drug interactions. Without this view, the model lacks the necessary granular details to accurately model molecular behavior, which leads to a significant drop in performance. Specifically, the atomic interactions  and molecular substructures are essential to understanding how a molecule behaves chemically, which is crucial for tasks that involve toxicity prediction or drug interaction forecasting. On the other hand, the drug view, while still important, does not provide the same atomic-level granularity. Instead, it focuses more on higher-level drug properties and interactions. While this information is useful for understanding a drug's broader functional characteristics, it is less essential for capturing the precise molecular behavior that influences properties like toxicity and binding affinity. Therefore, when the element view is retained, the drug view's removal does not lead to a severe performance drop for some tasks, as the model can still rely on the core atomic-level data provided by the element view.
\end{itemize}

\subsection*{Case Study}

\begin{figure*}[!ht]
  \centerline{\includegraphics[width=1\textwidth]{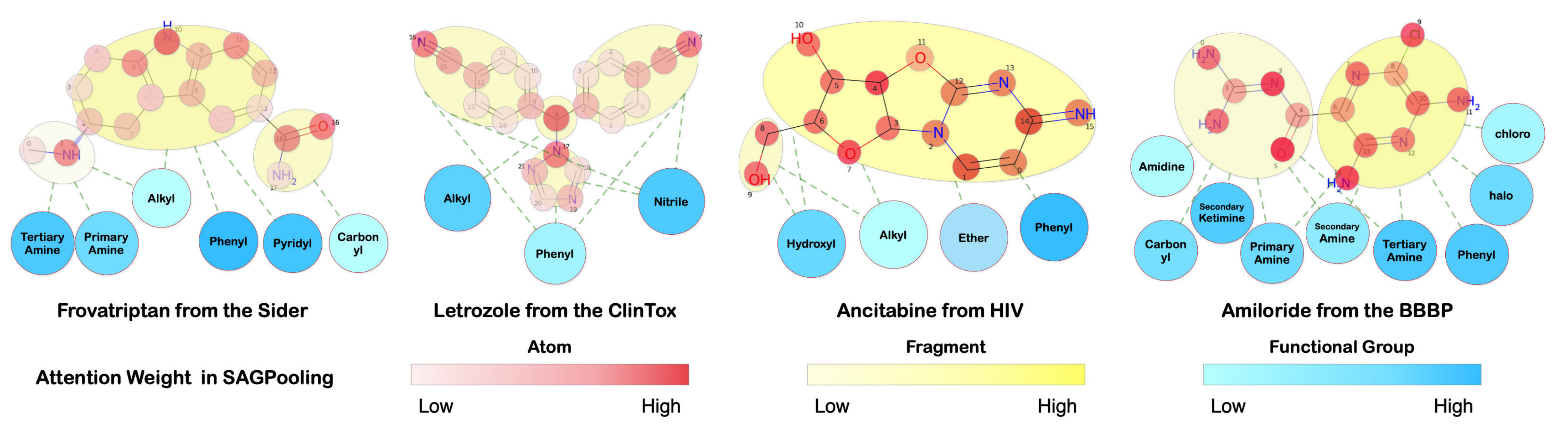}}
  \caption{Attention visualization. The weights are normalized across node types from the final SAGPolling layer, with darker colors denoting higher attention weights, thereby highlighting the areas of greater significance within the molecular structure.}
  \label{case}
\end{figure*}

The multi-view encoder approach provides multiple perspectives for interpreting molecular property predictions. This allows us not only to identify key structural elements but also to deepen our understanding of chemical properties and their impact on molecular behavior. The visualization of the attention weights across different molecular components is shown in Fig.\ref{case}. The weights are normalized across different node types from the final SAGPooling layer, with darker colors representing higher attention weights. This provides a clear indication of the regions of the molecular structure that are more significant in terms of the model’s predictive capability.

\paragraph*{Case Study 1: Frovatriptan from SIDER}
Frovatriptan, a medication used for migraine treatment, presents an interesting molecular structure, which is effectively analyzed by our multi-view encoder approach. Below is a detailed analysis of the attention weights assigned to various components of the molecule at the fragment level, atomic level, and functional group level.
\begin{itemize}
  \item Fragment Level: The \textbf{triptan ring}, a key structural component of Frovatriptan, receives the highest attention weight. This fragment is vital for the molecule’s binding to serotonin receptors, which is responsible for its therapeutic effect in treating migraines. The model recognizes the importance of this core structural unit and assigns it significant attention to facilitate the receptor interaction.
  \item Atomic Level: \textbf{The 1st and 10th nitrogen atoms} are assigned relatively high attention weights. These nitrogen atoms play a crucial role in the molecule’s interaction with the serotonin receptor. Their positioning in the molecule allows them to participate in the binding mechanism, making them essential for the drug's activity in the central nervous system. \textbf{The 15th carbon atom and 16th oxygen atom} also receive higher attention weights compared to other atoms. These atoms contribute to the overall stability of the molecule and its ability to interact with the biological targets.
  \item Functional Group Perspective: The \textbf{phenyl group} receives the highest attention weight among the functional groups. This group plays a significant role in enhancing the lipophilicity of the molecule, allowing it to cross biological membranes more easily and interact with the serotonin receptor more effectively. The \textbf{pyridyl group} also receives a high attention weight, indicating its importance in the drug’s pharmacological action. This group contributes to the molecule’s ability to target specific receptor sites, enhancing its therapeutic effect by providing the necessary interaction with serotonin receptors.
\end{itemize}

\paragraph*{Case Study 2: Ancitabine from the HIV Database}
The Ancitabine molecule, used for the treatment of HIV, is an analogue of Cytarabine that plays a crucial role in preventing viral replication. Our multi-view encoder approach, which integrates various molecular features, enables us to identify significant structural components and their corresponding influence on the molecule's behavior.
\begin{itemize}
  \item Fragment Level: The analysis highlights a crucial segment within the molecule’s structure—a \textbf{Cytarabine derivative}. This fragment, integral to the activity of Ancitabine, was assigned a higher attention weight. The encoder effectively identifies this key fragment as highly relevant to the molecule’s mechanism of action and importance in HIV treatment. The higher attention weight reflects the biological significance of this structural unit in inhibiting HIV replication.
  \item Atomic Level: A more granular investigation of atomic-level attention reveals varying importance levels for atoms within the Ancitabine molecule. Particularly, the \textbf{11th oxygen atom} is assigned a relatively low weight, which aligns with its functional role in \textbf{hydrolysis}. In the body, this oxygen atom plays a crucial role in the conversion of \textbf{Ancitabine to Cytarabine}, which prevents Thymidine from being incorporated into DNA, a key step in halting HIV replication. Despite its small proportion in the attention weight, this atom's function in the hydrolysis process is critical for the molecule’s efficacy.
  \item Functional Group Perspective: The attention analysis at the functional group level shows that the \textbf{Phenyl group} within Ancitabine was allocated the highest attention weight. This indicates that the Phenyl group has a significant role in the compound’s interaction with its biological target. The Alkyl structure, a common feature in many drug molecules, was given a lower weight, reflecting its relatively lesser importance in this particular molecule’s activity.
\end{itemize}

These findings underscore the ability of our HMG encoder to provide a layered and detailed exploration of molecular structures, enhancing predictive accuracy and offering valuable insights into molecular design and drug development.

\section*{Conclusion}
In this study, we introduced the cross-view KCHML method, which is a significant advancement in molecular property prediction.
We introduced and utilized HMG to integrate more external knowledge and capture finer molecular structure details.
KHCML's innovative multi-view framework and dual message passing mechanism provide a comprehensive molecular property prediction method that improves result accuracy and robustness. Furthermore, the method's effectiveness is demonstrated through extensive experiments, outperforming existing state-of-the-art methods.

In future work, we plan to further explore and expand on the use of heterogeneous graph-based methods for drug encoding. This will involve delving into how different types of data, including three-dimensional molecular structures and descriptive textual information, can be effectively integrated into the model. Such advancements could greatly enhance our understanding of drug properties and interactions, paving the way for more nuanced and precise drug discovery processes. 

However, a key limitation in our current approach is the reliance on available data sources, which may not always capture the full complexity of drug interactions or properties. To address this, we aim to incorporate more diverse and multimodal datasets, including clinical data and real-world evidence, to strengthen the model's generalization capabilities. Additionally, handling the cold-start problem, particularly in the inductive setting, remains a challenge that will require further investigation. We intend to explore strategies such as transfer learning or semi-supervised learning to mitigate these limitations and improve the model's performance with unseen drug pairs.

\nolinenumbers

\section*{Author contributions statement}
Chen and Hu conceived the experiments and analyzed the results, Chen experimented and wrote the manuscript. All authors guided the writing of the manuscript and reviewed the manuscript.

\section*{Financial Disclosure}
The work was supported by the National Key Research and Development Program of China (2023YFC2705700 to WH). This work was also supported in part by the Natural Science Foundation of China (No. 82174230 to WH), the Artificial Intelligence Innovation Project of the Wuhan Science and Technology Bureau (No. 2022010702040070 to WH), and the Natural Science Foundation of Shenzhen City (No. JCYJ20230807090211021 to WH). The funders had no role in study design, data collection and analysis, decision to publish, or preparation of the manuscript. The funders had no role in study design, data collection and analysis, decision to publish, or preparation of the manuscript.

\section*{Competing Interests}
No competing interest is declared. All authors have read and approved the final manuscript.

\section*{Supporting information}

\paragraph*{S1 Appendix.}
\label{S1_Appendix}
In the Appendix, we provide a detailed overview of work related to molecular representation learning and include additional experimental details. These include a description of the datasets, the initialization of the two knowledge graphs, feature extraction for nodes and edges in molecular modeling, and parameter selection details. We also present a theoretical proof demonstrating that the contrastive learning loss function effectively optimizes the model.

\end{document}